\documentclass[runningheads]{llncs}
\usepackage{graphicx}
\usepackage{CJKutf8}
\usepackage{color}
\usepackage{multicol}
\usepackage{multirow}
\usepackage{hyperref}

\begin{document}
\begin{CJK}{UTF8}{gkai}
\title{GrammarGPT: Exploring Open-Source LLMs for Native Chinese Grammatical Error Correction with Supervised Fine-Tuning}
\titlerunning{GrammarGPT}
%
\author{Yaxin Fan\inst{1,2,3} \and
Feng Jiang\inst{1,3,4}\thanks{Corresponding Author} \and
Peifeng Li\inst{2} \and Haizhou Li\inst{1,3}}

\authorrunning{Fan et al.}
%
\institute{School of Data Science, The Chinese University of Hong 
Kong, Shenzhen, China \and
School of Computer Science and Technology, Soochow University, China \and
Shenzhen Research Institute of Big Data, Shenzhen, Guangdong, China \and
School of Information Science and Technology, University of Science and Technology of China, China\\
\email{yxfansuda@stu.suda.edu.cn \\
pfli@suda.edu.cn \\
\{jeffreyjiang,haizhouli\}@cuhk.edu.cn}}

\maketitle              
\begin{abstract}
Grammatical error correction aims to correct ungrammatical sentences automatically. Recently, some work has demonstrated the excellent capabilities of closed-source Large Language Models (LLMs, e.g.,  ChatGPT) in grammatical error correction. However, the potential of open-source LLMs remains unexplored. In this paper, we introduced GrammarGPT, an open-source LLM, to preliminary explore its potential for native Chinese grammatical error correction. The core recipe of GrammarGPT is to leverage the hybrid dataset of ChatGPT-generated and human-annotated. For grammatical errors with clues, we proposed a heuristic method to guide ChatGPT to generate ungrammatical sentences by providing those clues. For grammatical errors without clues, we collected ungrammatical sentences from publicly available websites and manually corrected them. In addition, we employed an error-invariant augmentation method to enhance the ability of the model to correct native Chinese grammatical errors. We ultimately constructed about 1k parallel data and utilized these data to fine-tune open-source LLMs (e.g., Phoenix, released by The Chinese University of Hong Kong, Shenzhen) with instruction tuning. The experimental results show that GrammarGPT outperforms the existing SOTA system significantly. Although model parameters are 20x larger than the SOTA baseline, the required amount of data for instruction tuning is 1200x smaller, illustrating the potential of open-source LLMs on native CGEC. Our GrammarGPT ranks $3^{rd}$ on NLPCC2023 SharedTask1, demonstrating our approach's effectiveness. The code and data are available at \url{https://github.com/FreedomIntelligence/GrammarGPT}.

\keywords{Native Chinese grammatical error correction  \and Large language models \and ChatGPT \and Instruction tuning.}
\end{abstract}

\section{Introduction}
Grammatical Error Correction (GEC) aims to automatically correct ungrammatical sentences without changing their meaning \cite{zhang-etal-2022-mucgec,ma-etal-2022-linguistic,zhang2023nasgec}. Previous works \cite{zhao2018overview,rao2018overview,rao2020overview,zhang-etal-2022-mucgec} in Chinese Grammatical Error Correction (CGEC) mainly study the errors from foreign Chinese learners, which are very obvious and naive. Therefore, recent works ~\cite{zhang2023nasgec,ma-etal-2022-linguistic} shift to the grammatical errors made by native speakers, which are more subtle and challenging. Table~\ref{Examples} shows the six main types of grammatical errors made by native speakers, which can be divided into two types, e.g., with (w/) and without (w/o) clues. We can find that the incorrect sentences are fluent and in line with the habits of native Chinese. However, they do not conform to Chinese grammar, which is more difficult to correct.

Previous studies in GEC mainly adopted both Seq2edit \cite{hinson-etal-2020-heterogeneous,zhang-etal-2022-mucgec,liang-etal-2020-bert,ma-etal-2022-linguistic} and Seq2seq \cite{katsumata-komachi-2020-stronger,Zhao_Wang_2020,rothe-etal-2021-simple} paradigms and have achieved impressive performance on various GEC benchmarks.  With the emergence of LLMs, Fang et al. \cite{fang2023chatgpt} evaluated the performance of closed-source LLMs (e.g., ChatGPT \footnote{https://chat.openai.com/}) on GEC and revealed its excellent capabilities for error detection and correction. However, the potential of open-source LLMs remains unexplored.

In this paper, we introduce GrammarGPT, a novel model for studying the potential of open-source LLMs architectures in addressing Native Chinese Grammatical Error Correction (CGEC) through supervised fine-tuning. The key challenge in fine-tuning LLMs for CGEC is obtaining high-quality parallel data comprising grammatical errors made by native speakers. However, manually annotating such data is not only time-consuming but also expensive, necessitating the exploration of automatic data annotation methods. Recent works \cite{zhang2023huatuogpt,Taoli-LLama} have successfully leveraged distilled data from ChatGPT and real-world datasets to fine-tune LLMs for specific domains, effectively reducing costs while achieving superior performance. Inspired by this line of research, we propose a hybrid dataset that incorporates different types of native Chinese grammatical errors.

Specifically, we first proposed a heuristic method for the grammatical errors with clues as shown in Fig.~\ref{Examples} that guides ChatGPT to generate ungrammatical sentences by providing those clues. Then, for those errors without clues, we collected the ungrammatical sentences from the public website and corrected them manually. In addition, we proposed an error-invariant data augmentation method to enhance the diversity of the data by substituting the named entities in parallel data with similar ones, which can improve the ability of the model to correct native Chinese grammatical errors. We ultimately constructed 1k parallel data and utilized these data to fine-tune LLMs with instruction tuning. The experimental results show that GrammarGPT can significantly outperform state-of-the-art (SOTA) systems. Although the size of model parameters is 20x larger than the SOTA baseline, the data for fine-tuning is 1200x smaller, which demonstrated the potential of open-source LLMs on Chinese grammatical error correction.

\begin{table*}[t]\scriptsize
\caption{Examples of sentences with various types of grammatical errors. For those errors with clues, we can easily detect and correct them. For example, the co-occurrence of \emph{\textcolor{red}{超过}}(\emph{\textcolor{red}{more than}}) and \textcolor{green}{\emph{左右}} (\textcolor{green}{\emph{about}}) lead to redundant component error and we can remove one of them to make the sentence conform to Chinese grammar. However, for those errors without clues, a deeper understanding of Chinese grammar is required to detect and correct.} 
\label{Examples}
\centering
\centering
\begin{tabular}{ccl}
\hline
\multirow{3}{*}{ \begin{tabular}[l]{@{}c@{}} \\ \\ \\ \\ \\ w/ Clues\end{tabular}}     & \begin{tabular}[l]{@{}c@{}}Redundant \\ Component \\(RC)\end{tabular} & \begin{tabular}[l]{@{}l@{}}\textbf{Incorrect:}这座卫星城的人口估计 \textcolor{red}{超过}一百万\textcolor{green}{左右}。\\The population of this satellite city is estimated to be \\ \textcolor{red}{more than} \textcolor{green}{about} one million.\\ \textbf{Correct:}这座卫星城的人口估计超过一百万。\\The population of this satellite city is estimated to be \\over one million.\end{tabular}         \\\cline{2-3} 

& \begin{tabular}[l]{@{}c@{}}Structural \\ Confusion \\(SC)\end{tabular} & \begin{tabular}[l]{@{}l@{}}\textbf{Incorrect:}这次网络故障的\textcolor{red}{原因是由}服务器故障\textcolor{green}{引起的}。\\\textcolor{red}{The cause} of this network failure is \textcolor{green}{caused by} the server failure.\\ \textbf{Correct:}这次网络故障的原因是服务器故障。\\ The cause of the network failure is the server failure. \\\end{tabular}        \\\cline{2-3}

& \begin{tabular}[l]{@{}c@{}}Improper \\ Collocation \\(IC)\end{tabular} & \begin{tabular}[l]{@{}l@{}}\textbf{Incorrect:}西湖区正全面 \textcolor{red}{提升}区域产城融合发展的 \textcolor{green}{步伐}。\\Xihu District is \textcolor{red}{promoting} the \textcolor{green}{pace} of integration of regional \\industry and city development.\\\textbf{Correct:} 西湖区正全面加快区域产城融合发展的步伐。\\Xihu District is accelerating the pace of integration of regional \\ industry and city development.\end{tabular} \\ \hline                
                                                
\multicolumn{1}{c}{\multirow{3}{*}{ \begin{tabular}[l]{@{}c@{}} \\ \\ \\ \\ \\ w/o Clues\end{tabular}}}   & \begin{tabular}[l]{@{}c@{}}Improper\\ Word Order\\(IWO)\end{tabular}   & \begin{tabular}[l]{@{}l@{}}\textbf{Incorrect:}: 学校\textcolor{blue}{三个月内要求每名学生}完成20个小时的义工服务。\\ The school \textcolor{blue}{in three months requires each student} to complete \\20 hours of volunteer service.
\\\textbf{Correct:}学校要求每名学生三个月内完成20个小时的义工服务。\\The school requires each student to complete 20 hours of \\volunteer service in three months.\end{tabular}              \\ \cline{2-3} 
                                                & \begin{tabular}[l]{@{}c@{}}Improper\\ Logicality\\(IL)\end{tabular} & \begin{tabular}[l]{@{}l@{}}\textbf{Incorrect:}集团向社会各界人士、\textcolor{blue}{沿途村庄百姓}表示歉意。\\ The group apologizes to people from all walks of life \textcolor{blue}{and} \\ \textcolor{blue}{villagers along the way}.\\\textbf{Correct:}集团向社会各界人士表示歉意。\\ The group apologizes to people from all walks of life. \end{tabular}         \\ \cline{2-3} 

                                                & \begin{tabular}[l]{@{}c@{}}Missing \\ Component \\(MC)\end{tabular}  & \begin{tabular}[l]{@{}l@{}}\textbf{Incorrect:}这篇报告控诉了人类破坏大自然\textcolor{blue}{(...)}。\\The report accused man of destroying nature.\\ \textbf{Correct:}这篇报告控诉了人类破坏大自然的罪行。\\The report accused man the crime of destroying nature. \end{tabular}          \\\hline
\end{tabular}
\vspace{-0.8cm}
\end{table*}

Our contributions are as follows:
\begin{itemize}
    \item To the best of our knowledge, we are the first to explore the potential of open-source LLMs with instruction tuning for native Chinese grammatical error correction.
    \item We have constructed a hybrid dataset generated by ChatGPT and manual annotation, which can effectively cover native Chinese grammatical errors for taming the LLMs into an excellent grammar detector.
    \item We designed an error-invariant data augmentation method to substitute the named entities in parallel data with similar ones, making the model more accurate in correcting grammatical errors.
    \item The experimental results show that GrammarGPT can outperform the SOTA system significantly, and the data size for instruction tuning is only 1/1200 of the SOTA system.
\end{itemize}

\section{Related Work}
\subsection{Grammatical Error Correction}
The works in grammatical error correction can be divided into two paradigms: the Seq2edit paradigm and the Seq2seq paradigm.

\paragraph{Seq2edit paradigm} Seq2edit paradigm aims to predict the modification label, including insertion, deletion, and substitution, for each position of the sentence iteratively. Hinson et al. \cite{hinson-etal-2020-heterogeneous} proposed a heterogeneous approach to CGEC, composed of a NMT-based model, a sequence editing model, and a spell checker. Liang et al. \cite{liang-etal-2020-bert} introduced and transferred the BERT-fused NMT model and sequence tagging model into the CGEC field.  Zhang et al. \cite{zhang-etal-2022-mucgec} proposed a multi-reference multi-source evaluation dataset for CGEC and adopted the seq2edit method that enhanced with large pre-trained language models. Ma et al. \cite{ma-etal-2022-linguistic} propose a linguistic rules-based approach to construct large-scale CGEC training corpora with automatically generated grammatical errors and adopt the seq2edit method for evaluation.

\paragraph{Seq2seq paradigm} This paradigm treats CGEC as a monolingual translation task. Katsumata and Komachi \cite{katsumata-komachi-2020-stronger} explored the utility of bidirectional and auto-regressive transformers (BART) as a generic pre-trained encoder-decoder model for GEC. Zhao and Wang \cite{Zhao_Wang_2020} proposed a simple yet effective method to improve the NMT-based GEC models by dynamic masking, which can generate more diverse instances to enhance model generalization. Rothe et al. \cite{rothe-etal-2021-simple} proposed a language-agnostic method to generate a large number of synthetic examples, and then fine-tune large-scale multilingual language models.

In addition, several works \cite{liang-etal-2020-bert,hinson-etal-2020-heterogeneous,Li_Guo_Zhu_Sheng_Jiang_Ren_Xu_2022,zhang-etal-2022-mucgec} observe the complementary power of the above two paradigms, thus promoting the performance through the model ensemble. In this paper, we adopt the Se2seq paradigm to fine-tune LLMs with instruction tuning.

\subsection{Instruction Tuning for LLMs}
Instruction tuning \cite{wei2021finetuned,sanh2021multitask} can improve the ability of model generalization by learning from a large number of tasks guided by instruction, which has been successfully applied to fine-tune LLMs on some specific tasks. The work on task-specific instruction tuning can be categorized into three types by data sources: ChatGPT-generated, human-annotated, and hybrid dataset of ChatGPT and human.

\paragraph{ChatGPT-generated data} 
Several works adopted the data generated by ChatGPT to fine-tune LLMs in the form of instructions. Ho et al. \cite{ho2022large} proposed Fine-tune-CoT, a method that generates reasoning samples from LLMS to fine-tune smaller models, which enables substantial reasoning capability of small models.  Wang et al. \cite{wang2023scott} proposed SCOTT, a faithful knowledge distillation method to learn a small, self-consistent CoT model from a teacher model that is orders of magnitude.  Chen et al. \cite{chen2023efficient} explored distilling the reasoning ability of LLMs into a more compact student model for multimodal named entity and multimodal relation extraction. Chen et al. \cite{chen2023efficient} proposed a data synthesis framework built upon the data generation functions parameterized by LLMs and prompts and used synthesized data to fine-tune LLaMA.

\paragraph{Human-annotated data} 
Some works directly convert the supervised data into the format of instructions to fine-tune LLMs.
Zhang et al. \cite{zhang2023instruct} proposed to fine-tune LLaMA \cite{touvron2023llama} on financial sentiment analysis with a small portion of supervised financial sentiment analysis data. Wang et al. \cite{wang2023instructuie} proposed a unified information extraction framework based on instruction tuning to model various information extraction tasks and capture the inter-task dependency. 

\paragraph{Hybrid dataset of ChatGPT and human}
Recently, some works utilized the hybrid data of humans and ChatGPT/GPT-4 to fine-tune LLMs. Zhang et al. \cite{zhang2023huatuogpt} proposed to leverage both distilled data from ChatGPT and real-world data from doctors to fine-tune Bloom \cite{scao2022bloom}. 
Yu et al. \cite{Taoli-LLama} adopted a hybrid data of Chinese education and general-domain instructions \cite{peng2023instruction} generated by GPT-4  to fine-tune LLaMA \cite{touvron2023llama}.
In this paper, we follow this line and fine-tune LLMs on native CGEC with the hybrid dataset of ChatGPT-generated and human-annotated with instruction tuning.

\begin{figure}[t]
	\centering  
        \includegraphics[width=\linewidth]{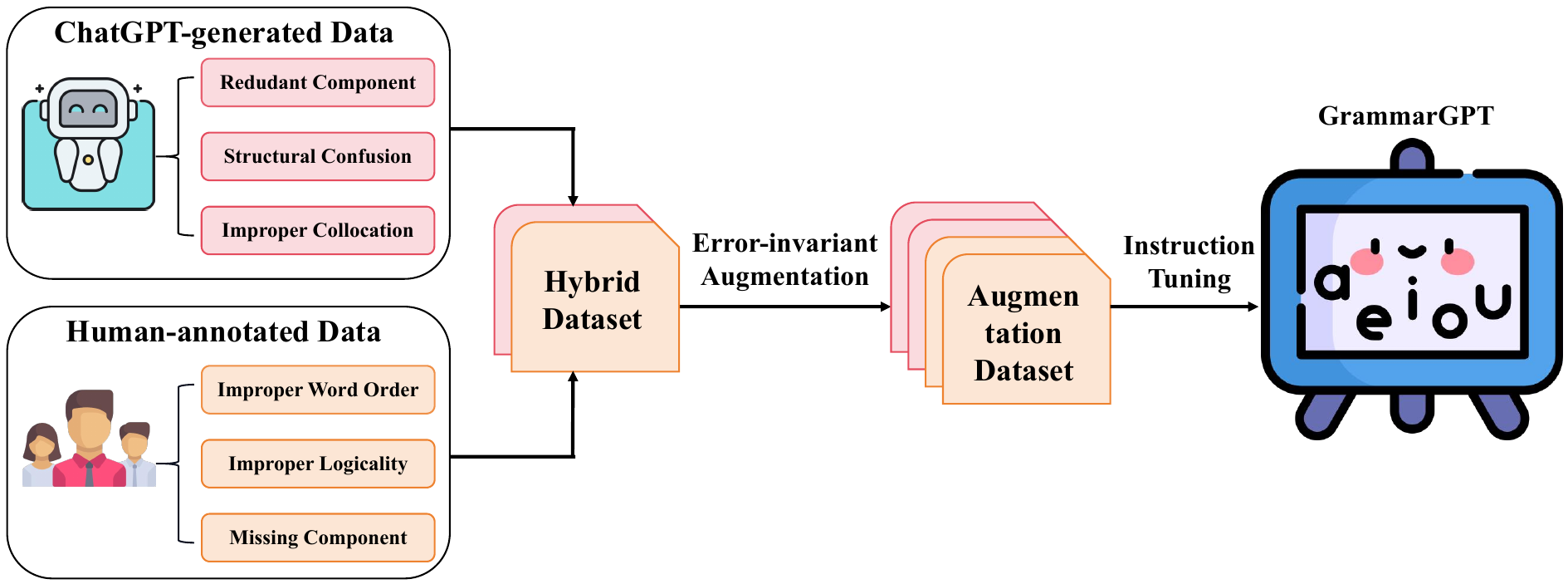}
        \caption{The framework of our method.}
         \label{Figure1}
          \vspace{-0.5cm}
    \end{figure}

\section{Methods}
Fig.~\ref{Figure1} illustrates the framework of our method, which involves the construction of parallel data comprising six types of native Chinese grammatical errors to facilitate the fine-tuning of open-source Language Model (LLMs). While human-annotated data offer high-quality samples, the associated high cost remains a significant concern. To address this, we adopt a compromise approach. We first guide ChatGPT to generate ungrammatical sentences with clues by providing those clues collected from the Internet. Then, we annotate the ungrammatical sentences without clues collected from the Internet. Additionally, we propose an error-invariant augmentation technique to substitute named entities in the parallel data with similar ones, further enhancing the model's capability to correct native Chinese grammatical errors. Finally, we convert the parallel data into instructions, which are then utilized for fine-tuning LLMs. Detailed explanations of these steps are provided in the following subsections.

\subsection{Hybrid Dataset Construction}
\subsubsection{ChatGPT-generated Data}
\begin{figure}[t]
\includegraphics[width=\textwidth]{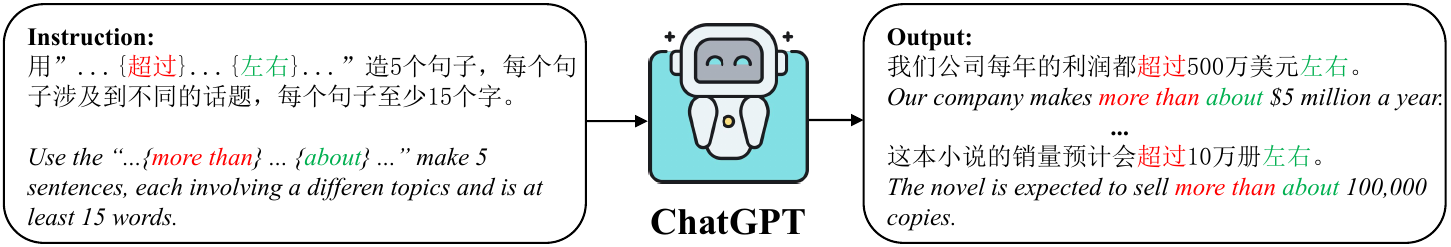}
\caption{Process of ungrammatical sentences generated by ChatGPT.} \label{GPTGenerate}
\vspace{-0.3cm}
\end{figure}

As shown in the first three lines of Table~\ref{Examples}, the grammatical errors with clues are easy to detect and correct by recognizing the specific clues. For example, \emph{"more than"} and \emph{"about"} are used together leading to \textbf{redundant component}, \emph{"The cause"} and \emph{"caused by"} are used together leading to \textbf{structural confusion}, and \emph{"prompting"} and \emph{"pace"} are used together leading to \textbf{improper collocation}. Conversely, we can construct the ungrammatical sentences by inserting these cues into grammatical sentences. Thanks to the strong capabilities of ChatGPT, we can instruct ChatGPT to generate the ungrammatical sentences that meet our requirements by providing these clues collected from public websites \footnote{https://wenku.baidu.com}. An example is as shown in Fig.~\ref{GPTGenerate}.  

\subsubsection{Human-annotated Data}
Some types of native ungrammatical errors are hard to recognize, as shown in the last three lines of Table~\ref{Examples}. We can find that those ungrammatical sentences are fluent and with no obvious clues of grammatical errors can help us to recognize them. For these types of grammatical errors, we mainly collected ungrammatical sentences from publicly available websites\footnote{https://tiku.baidu.com/} and then manually annotated them.

\begin{figure}[t]
\includegraphics[width=\textwidth]{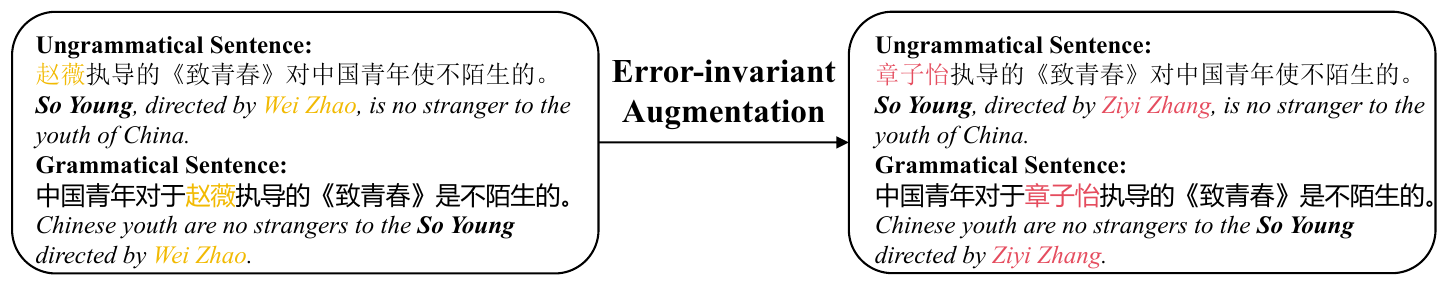}
\caption{An example of error-invariant augmentation.} \label{ErrorInvariantFigure}
\vspace{-0.5cm}
\end{figure}
\subsection{Error-invariant Data Augmentation}
To prioritize the model's focus on native grammar errors and improve its robustness, we have devised an error-invariant augmentation method,  as shown in Fig.~\ref{ErrorInvariantFigure}. Native Chinese grammatical errors are often subtle and infrequently found in the position of named entities. To address this, we adopt a strategy of substituting the named entities in the parallel data with similar ones\footnote{https://github.com/chatopera/Synonyms}. By employing this augmentation method, the model can concentrate on identifying unchanged errors rather than specific nouns, thereby improving its performance in correcting subtle and imperceptible grammar errors.

\subsection{Instruction Tuning}
Instruction tuning\cite{wei2021finetuned,sanh2021multitask} has emerged as the mainstream approach for fine-tuning LLMs by providing explicit instructions to enhance model comprehension. In this paper, we followed this mainstream trend and fine-tuned LLMs with instruction tuning. Instruction details are as shown in Table~\ref{instruction_details}, which mainly consists of four components.

1. \textbf{Task prefix}: This component guides LLMs to assume the role of an AI assistant.

2. \textbf{Task description}: Here, the specific task that LLMs are required to accomplish is outlined.

3. \textbf{Input}: This corresponds to ungrammatical sentences that are used as input during the fine-tuning process.

4. \textbf{Output}: This represents grammatical sentences, which serve as the expected output during fine-tuning.

\begin{table}[t]
\centering
\caption{Components of an instruction.}
\begin{tabular}{cl}
\hline
Instruction & \begin{tabular}[c]{@{}l@{}}\{\textbf{Task Prefix}\}\\\\
Human:\{\textbf{Task Description}\} \{\textbf{Input}\} Assistant :\{\textbf{Output}\}\end{tabular}\\
\hline
Task Prefix           & \begin{tabular}[c]{@{}l@{}}A chat between a curious human and an artificial \\ intelligence assistant. The assistant gives helpful, \\detailed, and polite answers to the human's questions.\end{tabular} \\
Task Description & Evaluate this sentence for grammar mistake                                                                                                                                                                 \\
Input         & \emph{Ungrammatical sentence}                                                                                                                                                                                \\
Output         & \emph{Grammatical sentence}       \\ \hline                                                                                                                                                                           
\end{tabular}
\label{instruction_details}
\vspace{-0.3cm}
\end{table}

\begin{table*}[t]
\centering
\caption{Statistic of the dataset.}
\begin{tabular}{c|c|ccc|ccc}
\hline
\multirow{3}{*}{Dataset} & \multirow{3}{*}{Number} & \multicolumn{6}{c}{Percentage of Different Grammatical Errors (\%)}           \\  
                      &                         & \multicolumn{3}{c|}{ChatGPT-generated}  & \multicolumn{3}{c}{Human-annotated} \\
                      &                         &  \quad RC  \quad         & \quad SC\quad           & \quad IC\quad           &IWO              &  \quad IL            & MC         \\ \hline
training set                 & 1061                   &  \quad23.54\quad &  \quad28.25\quad &  \quad13.70\quad & 6.50  &  \quad 13.18  & 15.07         \\ 
validating set                 & 500                     & \quad-          &\quad -          &\quad -          &\quad -            & \quad  -          & \quad-         \\ \hline
\end{tabular}
\label{statistic}
\vspace{-0.5cm}
\end{table*}

\section{Experiments}
\subsection{Datasets}
We constructed a total of 1061 parallel data samples for training, and the data statistics are provided in Table~\ref{statistic}. Roughly 35\% of the data were manually annotated, while the remaining 65\% were generated using ChatGPT. To evaluate the performance of our model, we utilized the validating set available on the NLPCC2023 SharedTask1 website\footnote{https://github.com/masr2000/NaCGEC}, which consists of 500 parallel data samples. We report the model's performance on this validating set for all the experiments conducted.

\subsection{Metrics}
The evaluation of a grammatical error correction system relies on the extent to which its proposed corrections or edits align with the gold-standard edits \cite{ng2014conll}. In line with previous research \cite{ma-etal-2022-linguistic,zhang-etal-2022-mucgec}, we adopt the word-level and char-level MaxMatch (M2) Scorer \cite{dahlmeier-ng-2012-better} for evaluation\footnote{https://github.com/HillZhang1999/MuCGEC/tree/main/scorers/ChERRANT}. This scorer computes Precision, Recall, and F$_{0.5}$ scores, comparing the gold edit set with the system edit set.

\subsection{Hyper-parameters}
The models are implemented in PyTorch using the Huggingface Transformers\footnote{https://huggingface.co/}. We used phoenix-inst-chat-7b \footnote{https://huggingface.co/FreedomIntelligence/phoenix-inst-chat-7b} \cite{chen2023phoenix} as the backbone. We set the max sequence length to 256. The model is trained with the AdamW optimizer, where the batch size and epoch are set to 64 and 3, respectively. We set the learning rate and the schedule type of learning rate to 2e-5 and 'linear', respectively. The warmup step is set to 5. The hyper-parameters are shown in Table~\ref{hyparameters_detail}.

\begin{table}[t]
\centering
\caption{Details of hyper-parameters.}
\begin{tabular}{c|c}
\hline
Backbone         & phoenix-inst-chat-7b \\ \hline
Max length       & 256     \\ \hline
Optimizer        & AdamW  \\ \hline
Batch size       & 64      \\ \hline
Epoch            & 1       \\ \hline
Learning rate    & 2e-5    \\ \hline
Lr schedule type & Linear  \\ \hline
Warmup steps     & 5       \\ \hline

\end{tabular}
\label{hyparameters_detail}
\vspace{-0.5cm}
\end{table}

\subsection{Experimental Results}
To validate the effectiveness of our method, we conducted a comparison between our GrammarGPT and the state-of-the-art (SOTA) baseline, S2S\_BART \cite{zhang-etal-2022-mucgec}. S2S\_BART utilizes Chinese BART as the pre-trained model and fine-tunes it on the Lang8 \cite{zhao2018overview} and HSK \cite{zhang2009features} datasets, which consist of approximately 1.2 million parallel data samples. We also fine-tuned S2S\_BART on the hybrid dataset that we constructed, and the results are presented in Table~\ref{performCom}.

Remarkably, we observed that S2S\_BART trained on our 1k hybrid dataset achieved 17.57 and 18.16 $F_{0.5}$ on Word-level and Char-level separately, which is comparable to that baseline model using the 1.2M data from foreign language speakers. We attribute this to the significant discrepancy between the grammatical errors made by foreign language speakers and native Chinese speakers, making it challenging to effectively improve the performance of native CGEC by relying solely on data from foreign language speakers. These results further highlight the effectiveness of our method in constructing a hybrid dataset that contains native Chinese grammatical errors.

Furthermore, our GrammarGPT exhibited substantial improvement with only about 1k data samples for fine-tuning, achieving 32.56 and 35.84 $F_{0.5}$, respectively. It is almost double the performance of baseline models, showcasing the remarkable potential of open-source LLMs in native CGEC. The final result on the official test set shows that our GrammarGPT ranks $3 ^ {rd} $~\footnote{https://github.com/masr2000/NaCGEC}.

\begin{table}[t]
\centering
\caption{Performance comparison between GrammarGPT and the SOTA baseline.}
\begin{tabular}{c|c|c|c|ccc|ccc}
\hline
\multirow{2}{*}{Model} & \multirow{2}{*}{\#Param.} & \multirow{2}{*}{Data} & \multirow{2}{*}{Data size} & \multicolumn{3}{c|}{Word-level}                                                    & \multicolumn{3}{c}{Char-level}                                                    \\
                       &                           &                       &                            & Prec                      & Rec                       & F$_{0.5}$                      & Prec                      & Rec                       & F$_{0.5}$                      \\ \hline
S2S\_BART              & 375M                      & \begin{tabular}[l]{@{}c@{}}Lang8\\HSK\end{tabular}             & 1.2M                       & 22.31                     & 10.14                     & 17.99                     & 22.13                     & 9.66                      & 17.59                     \\
S2S\_BART                 & 375M                      & Ours                  & 1061                        & 21.08                     & 10.54                    & 17.57                   & 22.09                     & 10.62
                    & 18.16  \\
GrammarGPT                   & 7B                        & Ours                  & 1061                       & \textbf{42.42}                     & \textbf{16.87}                     & \textbf{32.56}                   & \textbf{46.67}                     & \textbf{18.58}                     & \textbf{35.84}         \\ \hline           
\end{tabular}
\label{performCom}
\end{table}
\begin{table}[t]
\centering
\caption{Ablation study of our method.}\label{AblationStudyTable}
\begin{tabular}{cc|llllll}
\hline
\multicolumn{2}{c}{\multirow{2}{*}{Data}}             & \multicolumn{3}{|c|}{Word-level}                                                    & \multicolumn{3}{|c}{Char-level}                                                    \\
\multicolumn{2}{c}{}                                  & \multicolumn{1}{|c}{Prec}  & \multicolumn{1}{c}{Rec}   & \multicolumn{1}{c|}{F$_{0.5}$}  & \multicolumn{1}{c}{Prec}  & \multicolumn{1}{c}{Rec}   & \multicolumn{1}{c}{F$_{0.5}$}  \\ \hline
\multirow{3}{*}{w/o Augmentation} & \multicolumn{1}{|c|}{Human-annotated}   & 12.20                      & 1.51                      & \multicolumn{1}{c|}{5.04}                      & 13.89                     & 1.48                      & 5.19                      \\
                                  & \multicolumn{1}{|c|}{ChatGPT-generated} & 30.38                     & 7.21                      & \multicolumn{1}{c|}{18.49}                     & 30.86                     & 7.35                      & 18.83                     \\
                                  & \multicolumn{1}{|c|}{Hybrid dataset}             & 41.76                     & 11.45                     & \multicolumn{1}{c|}{27.30}                      & 44.32                     & 11.50                      & 28.22                     \\ \hline
\multirow{3}{*}{w/ Augmentation}  & \multicolumn{1}{|c|}{Human-annotated}   & 15.46                     & 4.52                      & \multicolumn{1}{c|}{10.42}                     & 16.48                     & 4.44                      & 10.68                     \\
                                  & \multicolumn{1}{|c|}{ChatGPT-generated} & 43.75                     & 6.33                      & \multicolumn{1}{c|}{20.04}                     & 44.90                      & 6.49                      & 20.56                     \\
                                  & \multicolumn{1}{|c|}{Hybrid dataset}             & \multicolumn{1}{c}{42.42} & \multicolumn{1}{c}{16.87} & \multicolumn{1}{c|}{32.56} & \multicolumn{1}{|c}{46.87} & \multicolumn{1}{c}{18.58} & \multicolumn{1}{c}{35.84} \\\hline
\end{tabular}
\vspace{-0.5cm}
\end{table}

\subsection{Ablation Study}

In our analysis of the impact of our contributions, namely the construction of a hybrid dataset and the error-invariant augmentation method, we present the results in Table~\ref{AblationStudyTable}.

Notably, the model trained on ChatGPT-generated data consistently outperforms that trained the human-annotated data, irrespective of whether data augmentation is applied. We attribute this observation to two primary reasons. First, the quantity of human-annotated data is smaller than the data generated by ChatGPT due to the high cost of human annotation. Second, grammatical errors without clues are more challenging to correct.

Additionally, our hybrid dataset demonstrates the potential for enhancing the performance of native CGEC. This finding substantiates the effectiveness of our approach in constructing the hybrid dataset consisting of native Chinese grammatical errors.

Moreover, by employing the error-invariant augmentation method, we observe our model trained on hybrid dataset has significant improvements in Recall and F$_{0.5}$ metrics but only minor improvements in Precision. It indicates that our augmentation technique enhances the model's ability to detect grammatical errors by forcing the model to pay more attention to grammar errors in the augmentation data.

\section{Conclusion}
In this paper, we introduce GrammarGPT, an open-source Large Language Model (LLM) specifically designed for native Chinese grammatical error correction. We first construct a hybrid dataset containing approximately 1k parallel data samples. It comprises both ChatGPT-generated data and human-annotated data for dealing with grammatical errors with and without clues. Additionally, we introduced an error-invariant augmentation method to improve the model's capabilities in native Chinese grammatical error correction by forcing the model to pay more attention to grammar errors in the augmentation data. We further fine-tune the open-source large-scale language model on the constructed dataset. Experimental results and in-depth analysis demonstrate the effectiveness of our GrammarGPT in native Chinese grammatical error correction.

\section*{Acknowledge}
This work is supported by the National Natural Science Foundation of China (Grant No. 62271432) and the Guangdong Provincial Key Laboratory of Big Data Computing, The Chinese University of Hong Kong, Shenzhen (Grant No. B10120210117).

%
%
%
\bibliographystyle{splncs04}
\bibliography{splncs04}

\begin{thebibliography}{10}
\providecommand{\url}[1]{\texttt{#1}}
\providecommand{\urlprefix}{URL }
\providecommand{\doi}[1]{https://doi.org/#1}

\bibitem{chen2023efficient}
Chen, F., Feng, Y.: {Chain-of-Thought Prompt Distillation for Multimodal Named
  Entity and Multimodal Relation Extraction}. ArXiv preprint arXiv:2306.14122
  (2023)

\bibitem{chen2023phoenix}
Chen, Z., Jiang, F., Chen, J., Wang, T., Yu, F., Chen, G., Zhang, H., Liang,
  J., Zhang, C., Zhang, Z., et~al.: {Phoenix: Democratizing ChatGPT across
  languages}. arXiv preprint arXiv:2304.10453  (2023)

\bibitem{dahlmeier-ng-2012-better}
Dahlmeier, D., Ng, H.T.: {Better Evaluation for Grammatical Error Correction}.
  In: Proceedings of the 2012 Conference of the North {A}merican Chapter of the
  Association for Computational Linguistics: Human Language Technologies. pp.
  568--572. Association for Computational Linguistics, Montr{\'e}al, Canada
  (Jun 2012)

\bibitem{fang2023chatgpt}
Fang, T., Yang, S., Lan, K., Wong, D.F., Hu, J., Chao, L.S., Zhang, Y.: {Is
  ChatGPT a Highly Fluent Grammatical Error Correction System? A Comprehensive
  Evaluation}. arXiv preprint arXiv:2304.01746  (2023)

\bibitem{hinson-etal-2020-heterogeneous}
Hinson, C., Huang, H.H., Chen, H.H.: {Heterogeneous Recycle Generation for
  {C}hinese Grammatical Error Correction}. In: Proceedings of the 28th
  International Conference on Computational Linguistics. pp. 2191--2201 (2020)

\bibitem{ho2022large}
Ho, N., Schmid, L., Yun, S.Y.: {Large Language Models Are Reasoning Teachers}.
  arXiv preprint arXiv:2212.10071  (2022)

\bibitem{katsumata-komachi-2020-stronger}
Katsumata, S., Komachi, M.: {Stronger Baselines for Grammatical Error
  Correction Using a Pretrained Encoder-Decoder Model}. In: Proceedings of the
  1st Conference of the Asia-Pacific Chapter of the Association for
  Computational Linguistics and the 10th International Joint Conference on
  Natural Language Processing. pp. 827--832 (2020)

\bibitem{Li_Guo_Zhu_Sheng_Jiang_Ren_Xu_2022}
Li, J., Guo, J., Zhu, Y., Sheng, X., Jiang, D., Ren, B., Xu, L.:
  {Sequence-to-Action: Grammatical Error Correction with Action Guided Sequence
  Generation}. Proceedings of the AAAI Conference on Artificial Intelligence
  \textbf{36}(10),  10974--10982 (2022)

\bibitem{liang-etal-2020-bert}
Liang, D., Zheng, C., Guo, L., Cui, X., Xiong, X., Rong, H., Dong, J.: {{BERT}
  Enhanced Neural Machine Translation and Sequence Tagging Model for {C}hinese
  Grammatical Error Diagnosis}. In: Proceedings of the 6th Workshop on Natural
  Language Processing Techniques for Educational Applications. pp. 57--66.
  Association for Computational Linguistics (2020)

\bibitem{ma-etal-2022-linguistic}
Ma, S., Li, Y., Sun, R., Zhou, Q., Huang, S., Zhang, D., Yangning, L., Liu, R.,
  Li, Z., Cao, Y., Zheng, H., Shen, Y.: {Linguistic Rules-Based Corpus
  Generation for Native {C}hinese Grammatical Error Correction}. In: Findings
  of the Association for Computational Linguistics: EMNLP 2022. pp. 576--589
  (2022)

\bibitem{ng2014conll}
Ng, H.T., Wu, S.M., Briscoe, T., Hadiwinoto, C., Susanto, R.H., Bryant, C.:
  {The CoNLL-2014 Shared Task on Grammatical Error Correction}. In: Proceedings
  of the Eighteenth Conference on Computational Natural Language Learning:
  Shared Task. pp. 1--14 (2014)

\bibitem{peng2023instruction}
Peng, B., Li, C., He, P., Galley, M., Gao, J.: {Instruction Tuning with GPT-4}.
  arXiv preprint arXiv:2304.03277  (2023)

\bibitem{rao2018overview}
Rao, G., Gong, Q., Zhang, B., Xun, E.: {Overview of NLPTEA-2018 Share Task
  Chinese Grammatical Error Diagnosis}. In: Proceedings of the 5th Workshop on
  Natural Language Processing Techniques for Educational Applications. pp.
  42--51 (2018)

\bibitem{rao2020overview}
Rao, G., Yang, E., Zhang, B.: {Overview of NLPTEA-2020 Shared Task for Chinese
  grammatical error diagnosis}. In: Proceedings of the 6th Workshop on Natural
  Language Processing Techniques for Educational Applications. pp. 25--35
  (2020)

\bibitem{rothe-etal-2021-simple}
Rothe, S., Mallinson, J., Malmi, E., Krause, S., Severyn, A.: {A Simple Recipe
  for Multilingual Grammatical Error Correction}. In: Proceedings of the 59th
  Annual Meeting of the Association for Computational Linguistics and the 11th
  International Joint Conference on Natural Language Processing (Volume 2:
  Short Papers). pp. 702--707 (2021)

\bibitem{sanh2021multitask}
Sanh, V., Webson, A., Raffel, C., Bach, S.H., Sutawika, L., Alyafeai, Z.,
  Chaffin, A., Stiegler, A., Scao, T.L., Raja, A., et~al.: {Multitask Prompted
  Training Enables Zero-shot Task Generalization}. arXiv preprint
  arXiv:2110.08207  (2021)

\bibitem{scao2022bloom}
Scao, T.L., Fan, A., Akiki, C., Pavlick, E., Ili{\'c}, S., Hesslow, D.,
  Castagn{\'e}, R., Luccioni, A.S., Yvon, F., Gall{\'e}, M., et~al.: {Bloom: A
  176B-parameter Open-access Multilingual Language Model}. arXiv preprint
  arXiv:2211.05100  (2022)

\bibitem{touvron2023llama}
Touvron, H., Lavril, T., Izacard, G., Martinet, X., Lachaux, M.A., Lacroix, T.,
  Rozière, B., Goyal, N., Hambro, E., Azhar, F., Rodriguez, A., Joulin, A.,
  Grave, E., Lample, G.: {LLaMA: Open and Efficient Foundation Language Models}
  (2023)

\bibitem{wang2023scott}
Wang, P., Wang, Z., Li, Z., Gao, Y., Yin, B., Ren, X.: {SCOTT: Self-Consistent
  Chain-of-Thought Distillation}. arXiv preprint arXiv:2305.01879  (2023)

\bibitem{wang2023instructuie}
Wang, X., Zhou, W., Zu, C., Xia, H., Chen, T., Zhang, Y., Zheng, R., Ye, J.,
  Zhang, Q., Gui, T., et~al.: {InstructUIE: Multi-task Instruction Tuning for
  Unified Information Extraction}. arXiv preprint arXiv:2304.08085  (2023)

\bibitem{wei2021finetuned}
Wei, J., Bosma, M., Zhao, V.Y., Guu, K., Yu, A.W., Lester, B., Du, N., Dai,
  A.M., Le, Q.V.: {Finetuned Language Models Are Zero-shot Learners}. arXiv
  preprint arXiv:2109.01652  (2021)

\bibitem{Taoli-LLama}
Yu, J., Zhu, J., Wang, Y., Liu, Y., Chang, H., Nie, J., Kong, C., Cong, R.,
  XinLiu, An, J., Lu, L., Fang, M., Zhu, L.: {Taoli LLaMA}.
  \url{https://github.com/blcuicall/taoli} (2023)

\bibitem{zhang2009features}
Zhang, B.: {Features and Functions of the HSK Dynamic Composition Corpus}.
  International Chinese Language Education  \textbf{4},  71--79 (2009)

\bibitem{zhang2023instruct}
Zhang, B., Yang, H., Liu, X.Y.: {Instruct-FinGPT: Financial Sentiment Analysis
  by Instruction Tuning of General-Purpose Large Language Models}. arXiv
  preprint arXiv:2306.12659  (2023)

\bibitem{zhang2023huatuogpt}
Zhang, H., Chen, J., Jiang, F., Yu, F., Chen, Z., Li, J., Chen, G., Wu, X.,
  Zhang, Z., Xiao, Q., et~al.: {HuatuoGPT, towards Taming Language Model to Be
  a Doctor}. arXiv preprint arXiv:2305.15075  (2023)

\bibitem{zhang-etal-2022-mucgec}
Zhang, Y., Li, Z., Bao, Z., Li, J., Zhang, B., Li, C., Huang, F., Zhang, M.:
  {{M}u{CGEC}: a Multi-Reference Multi-Source Evaluation Dataset for {C}hinese
  Grammatical Error Correction}. In: Proceedings of the 2022 Conference of the
  North American Chapter of the Association for Computational Linguistics:
  Human Language Technologies. pp. 3118--3130 (2022)

\bibitem{zhang2023nasgec}
Zhang, Y., Zhang, B., Jiang, H., Li, Z., Li, C., Huang, F., Zhang, M.: {NaSGEC:
  a Multi-Domain Chinese Grammatical Error Correction Dataset from Native
  Speaker Texts}. arXiv preprint arXiv:2305.16023  (2023)

\bibitem{zhao2018overview}
Zhao, Y., Jiang, N., Sun, W., Wan, X.: {Overview of the NLPCC 2018 Shared Task:
  Grammatical Error Correction}. In: Natural Language Processing and Chinese
  Computing: 7th CCF International Conference, NLPCC 2018, Hohhot, China,
  August 26--30, 2018, Proceedings, Part II 7. pp. 439--445. Springer (2018)

\bibitem{Zhao_Wang_2020}
Zhao, Z., Wang, H.: {{MaskGEC}: Improving Neural Grammatical Error Correction
  via Dynamic Masking}. Proceedings of the AAAI Conference on Artificial
  Intelligence  \textbf{34}(01),  1226--1233 (2020)

\end{thebibliography}

%




\end{CJK}
\end{document}